\documentclass{article}
\usepackage{amsmath}

 \usepackage[preprint]{neurips_2026}

\usepackage[utf8]{inputenc} % allow utf-8 input
\usepackage[T1]{fontenc}    % use 8-bit T1 fonts
\usepackage{hyperref}       % hyperlinks
\usepackage{url}            % simple URL typesetting
\usepackage{booktabs}       % professional-quality tables
\usepackage{graphicx}
\usepackage{amsfonts}       % blackboard math symbols
\usepackage{nicefrac}       % compact symbols for 1/2, etc.
\usepackage{microtype}      % microtypography
\usepackage{xcolor}         % colors
\usepackage{verbatim}
\title{Reward Basis Decomposition for Interpreting Preference Heterogeneity}

\author{%
  Mohamed I. Eldagla\thanks{Equal contribution.} \\
  German International University in Cairo \\
  \texttt{mohamed.eldagla@giu-uni.de}
  \And
  Ifesi Onubogu\footnotemark[1] \\
  Columbia University \\
  \texttt{ifesi.onubogu@columbia.edu}
  \And
  Prerana Rane\footnotemark[1] \\
  PRISM AI Safety Research Fellow \\
  \texttt{preranarane@ieee.org}
  \And
  Harsh Sharma \\
  Carnegie Mellon University \\
  \texttt{harshsharma@cmu.edu}
  \And
  Rachel Freedman \\
  University of California, Berkeley \\
  \texttt{rachel.freedman@berkeley.edu}
}

\usepackage{xcolor}
\newcommand{\rev}[1]{#1}
\begin{document}

\maketitle

\begin{abstract}
Canonical reward models specify the optimization target for AI policies, so
understanding the values they encode is important for ensuring AI alignment.
Existing interpretability techniques interpret reward models with a single
reward head, which cannot distinguish between the values that a population
shares and those that vary among individuals. We introduce Reward Basis
Decomposition (RBD), which splits per-user weights into a population term and
a per-user term. This yields a shared population direction and individual user
directions that are analyzed separately via concept alignment and sparse
autoencoder decomposition. \iffalse We verify that RBD can recover artificially planted
per-user directions on a synthetic control, then apply it to two real-world
multi-preference datasets, PRISM and Community Alignment. In both, we find no
individual-user structure, suggesting that a single direction in the reward embedding space accounts for most preferences. \fi 
\rev{We verify that RBD can recover artificially planted per-user
directions on a synthetic control, then apply it to two real-world
multi-preference datasets, PRISM and Community Alignment. In both,
we find no reliable predictive benefit from the fitted per-user
terms beyond the shared population direction under our setup.} Code and the synthetic dataset are available at
 \url{https://github.com/RachelFreedman/RBD}.
\end{abstract}

\section{Introduction}

% What is the problem?
Reinforcement learning from human feedback (RLHF) aligns large language models by
optimizing a policy against a reward model trained on pairwise
preferences~\cite{ouyang2022traininglanguagemodelsfollow}, so understanding what that
model encodes tells a practitioner what behavior will be incentivized. Recent
mechanistic interpretability work~\cite{nadaf2026rewardlens, shi2025safer, christian2025reward,
laguna2025cbrm} interprets reward models with a single reward
head, which inherently collapses diverse users' preferences.

% Why is studying it important?
Preference datasets such as PRISM~\cite{kirk2024prism} and Community
Alignment~\cite{zhang2026community} preserve annotator disagreement by recording per-user
labels. Personalization methods such as Low-Rank Reward Modeling (LoRe)~\cite{bose2025lorepersonalizingllmslowrank} decompose reward into a shared basis with per-user weights to fit these datasets. These methods have a per-user structure that monolithic reward models lack. However, a single weight vector per user
leaves population and individual content entangled, making neither interpretable on its own.

% What did you do?
Reward Basis Decomposition (RBD, Section~\ref{sec:method}) addresses both limitations by
estimating a
small set of basis reward functions with per-user weights, then factoring those weights into a shared population term and a per-user deviation.
Weights are unconstrained in sign, allowing users to reward or penalize the same basis
component. 
\iffalse We evaluate RBD against LoRe on PRISM and Community
Alignment (Section~\ref{sec:conditions}) and include two synthetic controls in which per-user structure is
present by construction to distinguish the absence of structure in the data from method failure (Sections~\ref{sec:synthetic}
and~\ref{sec:perturbation}). \fi
\rev{We evaluate RBD against LoRe on PRISM and Community
Alignment (Section~\ref{sec:conditions}) and include two synthetic
controls in which per-user structure is present by construction
to test whether the method can recover known forms of per-user
structure (Sections~\ref{sec:synthetic}
and~\ref{sec:perturbation}).}

% What is your key insight?
We separate the population term from the per-user terms, allowing each component to be evaluated and analyzed independently. On synthetic data with planted per-user directions, a user's own
direction outperforms another user's by $+0.40$ in held-out accuracy, against
approximately $0.00$ on PRISM and Community Alignment. \iffalse Injecting a preference
signal into PRISM embeddings raises held-out accuracy from $0.54$
to $0.83$, confirming that the null result reflects the absence of structure rather than
sparse data. \fi \rev{Injecting a preference signal into PRISM embeddings raises
held-out accuracy from $0.54$ to $0.83$, showing that the same
per-user data budget permits recovery of the injected signal,
without ruling out weaker or differently encoded individual
preferences in the original data.} Our contributions are as follows:

\begin{itemize}
\item A method separating the shared population direction from the
individual user directions in a multi-user preference dataset, exposing both for independent interpretation.
\item An application to PRISM and Community Alignment showing their disagreement is not recoverable as per-user reward directions, and a
characterization of the shared direction that carries the model's accuracy.
\item A measurement showing concept directions depend on the model generating their contrastive text, at a magnitude exceeding the alignments
being measured.
\end{itemize}

% Why are your results important?
RBD introduces a new method for interpreting reward models. The population direction is distinct from the base reward model, stable, and more predictive of held-out
pairs (Section~\ref{sec:sae}). Concept alignment and sparse autoencoder (SAE)
decomposition apply directly to it using existing tooling. We hope to facilitate future work on interpreting reward models that represent diverse populations.

\section{Related Work}

\paragraph{Personalized reward modeling and preference datasets}
LoRe~\cite{bose2025lorepersonalizingllmslowrank} is a method for reward model personalization that factors per-user reward through
a low-rank basis shared across users. RBD builds on that factorization, further separating population and individual-level components, and improving upon LoRe's predictive accuracy (see Figure~\ref{fig:rank-panels}).
We use two large, open-source preference datasets that record annotator disagreement:
PRISM~\cite{kirk2024prism}, with feedback from 1,500 participants across 75
countries on value-laden prompts, and Community
Alignment~\cite{zhang2026community}, a five-country, 15,000-participant study.

\paragraph{Reward model interpretability}
Reward-lens~\cite{nadaf2026rewardlens} adapts logit-lens analogues, activation
patching, and SAE feature alignment to reward models by analyzing the reward head's
weight vector. SAFER~\cite{shi2025safer} trains an SAE on reward model activations and compares features between chosen and
rejected responses.
\citet{christian2025reward} treat reward models as black boxes and score every
vocabulary token across ten open-source reward models, serving as a behavioral fallback
when geometric concept alignment is inconclusive.

\paragraph{Concept and persona directions in representation space}
RBD's interpretability approach rests on the premise that a model's
residual stream decomposes into interpretable linear directions.
\citet{bricken2023monosemanticity} established that SAEs recover sparse, monosemantic features, and
\citet{chen2025personavectors} show that the mean activation difference between
contrastively prompted responses yields persona vectors
causally sufficient for steering behavior.

\section{Methods}
\label{sec:method}

\subsection{Background: low-rank reward personalization}

Reward models that serve many users must represent both common and diverging preferences. Our method factors out the common and diverging elements and analyzes them individually.

LoRe~\cite{bose2025lorepersonalizingllmslowrank} personalizes a reward model by factoring
per-user reward functions through a basis shared across users. A frozen backbone maps a
prompt $x$ and a response $y$ to an embedding $e(x,y) \in \mathbb{R}^d$, with $d$
the backbone's hidden width. The method learns a basis $V \in \mathbb{R}^{d \times K}$ whose
$K$ columns $V_{:,1}, \dots, V_{:,K}$ are candidate reward directions in that embedding
space, and assigns each user $u$ a weight vector $w_u \in \mathbb{R}^{K}$ that records how much
that user cares about each direction. User $u$'s reward is defined so all users share $V$ and differ only through $w_u$.
\begin{equation}
r_u(x,y) = e(x,y)^\top V w_u ,
\label{eq:reward}
\end{equation}

Training uses pairwise preference data. For a pair in which response $y^+$ was preferred to
$y^-$, write the embedding difference $\Delta = e(x,y^+) - e(x,y^-)$. A Bradley--Terry model
turns a difference in reward into a probability of preference. Hence, the probability that user
$u$ prefers $y^+$ is $\sigma(\Delta^\top V w_u)$, where $\sigma$ is the logistic function,
and fitting maximizes $\log \sigma(\Delta^\top V w_u)$ summed over that user's pairs.

LoRe constrains each $w_u$ to the probability simplex, via a softmax reparameterization,
\begin{equation}
w_u \in \mathcal{S}^{K-1} = \Big\{ w \in \mathbb{R}^K : w_k \geq 0 \ \ \forall k, \ \
\textstyle\sum_{k=1}^{K} w_k = 1 \Big\},
\label{eq:simplex}
\end{equation}

Both LoRe and RBD use a pretrained \textit{base reward model} ("base RM") to embed responses in reward space. We call this embedding the \emph{reward head} and write it as $v_{\mathrm{sft}}$. The goal of reward model personalization techniques like LoRe and RBD is to improve on the base RM's predictive accuracy for datasets of diverse preferences like PRISM and Community Alignment. LoRe regularizes by penalizing divergence from $v_{\mathrm{sft}}$, so we refer to $v_{\mathrm{sft}}$ as an \textit{anchor}.

% $V$ is regularized toward that vector, penalizing $1 - \cos(V_{:,k},
% v_{\mathrm{sft}})$ with weight $\alpha \geq 0$. The penalty is in the code we fork and run
% as our baseline, but not in the paper, whose training objective is the logistic loss alone.
% We refer to it as an \emph{anchor}: it pulls every learned basis column toward
% $v_{\mathrm{sft}}$, so larger $\alpha$ trades the freedom to learn new directions against
% reproducing the pretrained reward model. In the limit, every column is parallel to
% $v_{\mathrm{sft}}$. Since the weights are non-negative, every user's reward is then a
% positive multiple of the pretrained one and ranks responses identically to it. At $\alpha =
% 0$, the penalty vanishes, and the objective matches the paper.

\subsection{Reward Basis Decomposition}
\label{sec:rbd}
 \begin{figure}[htb]
  \centering
  \includegraphics[width=\linewidth]{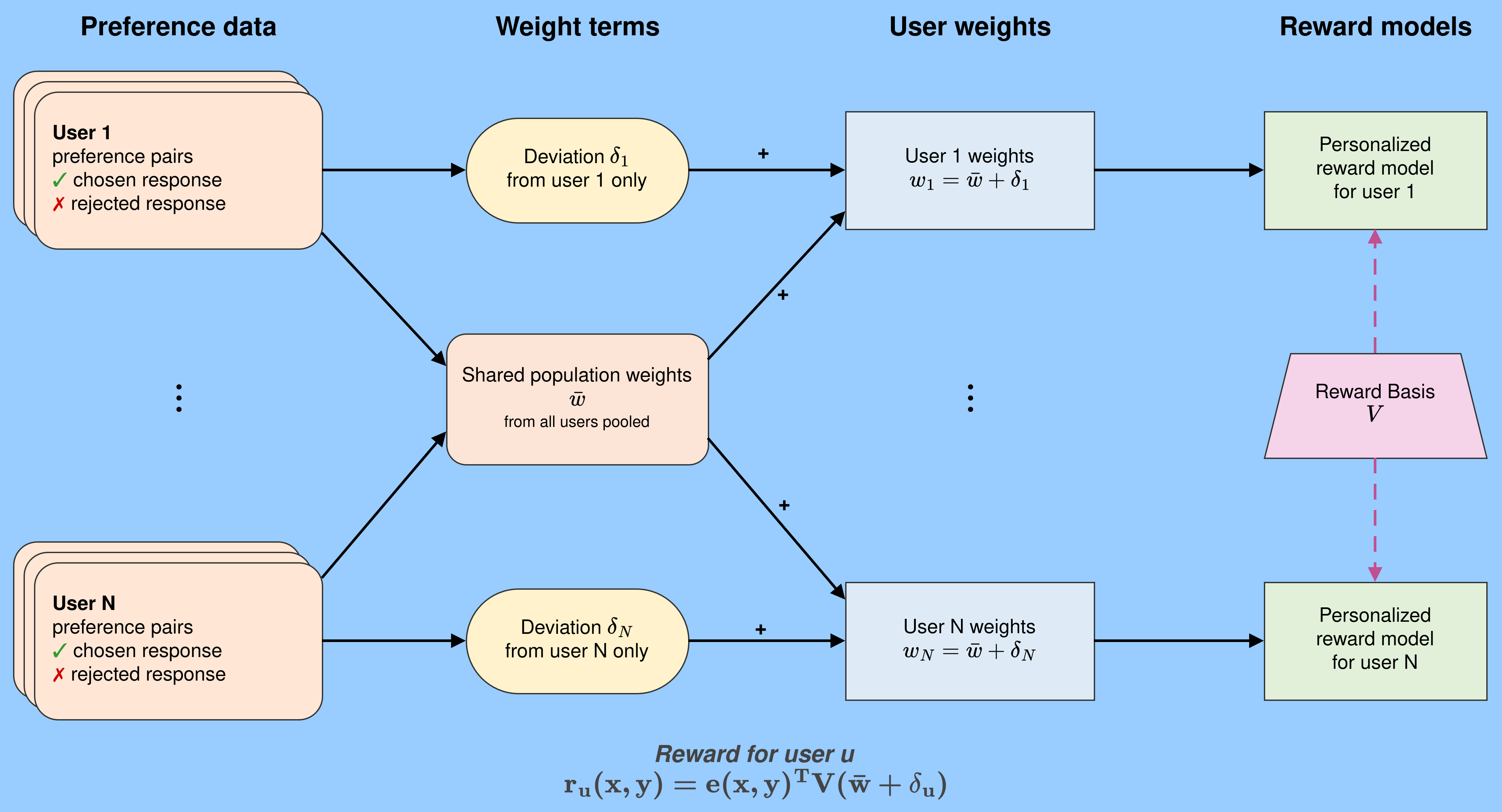}
  \caption{\textbf{Reward Basis Decomposition (RBD).} Each user's weights split into a population term $\bar{w}$, fit to the shared population preferences, and a deviation $\delta_u$, fit to user $u$'s pairs alone, so $w_u = \bar{w} + \delta_u$. All users share one reward basis $V$ (dashed), and user $u$'s reward is $r_u(x,y) = e(x,y)^\top V(\bar{w} + \delta_u)$. This lets the shared direction $V\bar{w}$ and the per-user directions $V(\bar{w} + \delta_u)$ be analyzed separately.}
  \vspace{-5pt}
  \label{fig:rbd-overview}
\end{figure}

\textit{Reward Basis Decomposition} (RBD) is a method for isolating and
interpreting the preference structure a multi-user reward model has learned. RBD fits Equation~\ref{eq:reward} where each user's weights are $w_u = \bar{w} +
\delta_u$. $\bar{w}$ is shared across the population and $\delta_u$ is specific to user $u$. $V\bar{w}$ is therefore a \textit{population direction} in reward embedding space, describing preferences common to the population, and $V(\bar{w} + \delta_u)$ is the \textit{user direction} for user $u$, describing how that user departs from it. Because the
split is explicit, the two parts can be evaluated and interpreted separately:
setting every $\delta_u = 0$ leaves a single population reward model, and
comparing it with the full model measures what the per-user terms contribute.

The weights are unconstrained in sign, so two users can reward and penalize the
same basis direction. A user who prefers what another dislikes is representable even at $K=1$ (LoRe's weights are restricted to the simplex of
Equation~\ref{eq:simplex}).

\paragraph{Objective} RBD maximizes the likelihood pooled over
all pairs, with ridge penalties on the magnitude of the population and per-user
reward directions:
\begin{equation}
  \mathcal{R} = \lambda_{\mathrm{pop}} \|V\bar{w}\|^2
    + \lambda_\delta \frac{1}{N}\sum_{u=1}^{N} \|V\delta_u\|^2 ,
  \label{eq:ridge}
\end{equation}
\begin{equation}
  \mathcal{L} = -\frac{1}{|\mathcal{D}|}\sum_u \sum_{\Delta\in\mathcal{D}_u}
    \log\sigma\!\big(\Delta^\top V(\bar{w}+\delta_u)\big) + \mathcal{R},
  \label{eq:objective}
\end{equation}
where $N$ is the number of users, $\mathcal{D}_u$ is user $u$'s set of pairs, and
$\mathcal{D}$ is their union. Pooling means users contribute in proportion to
their data. $\lambda_{\mathrm{pop}}$ and $\lambda_\delta$ control the shared
reward and the deviations separately and independently of rank. The
penalty is on magnitude rather than on divergence from $v_{\mathrm{sft}}$. All $K$ columns are retained.

\paragraph{Optimization} $V$, $\bar{w}$ and all $\delta_u$ are optimized jointly, with no fixed rescaling of the logits. In our implementation $d=4096$, so the scale of $V$ at initialization is set from the data so that initial
logits have unit standard deviation; otherwise the sigmoid saturates and
gradients vanish. Training stops when the validation objective plateaus.

\paragraph{Tuning protocol}
$\lambda_{\mathrm{pop}}$, $\lambda_{\delta}$ and the learning rate are selected on a
held-out validation split, never on test, at fixed $K = 8$, and the selected values are
reused at every rank so that regularization strength does not depend on $K$. The sweep is given in
Appendix~\ref{app:tuning}.

\paragraph{Identifiability}
\label{sec:identifiability}
Removing the simplex constraint costs \emph{identifiability}: for any invertible
$R \in \mathbb{R}^{K \times K}$, $V \mapsto VR$, $w \mapsto R^{-1}w$ leaves the objective
unchanged (Appendix~\ref{app:identifiability}), so the column space of $V$ and the induced
reward functions are identified, but individual columns are not. We report the shared direction $V\bar{w}$ and the per-user directions $V(\bar{w}+\delta_u)$ which are transform-invariant quantities.

\paragraph{Synthetic control with planted personalization}
\label{sec:synthetic}
To verify our method, we synthesize a dataset with known per-user structure, and confirm that RBD can recover it. Each synthetic user is a \emph{persona}, a sparse signed weighting over concept axes, and all users judge the same responses (see Appendix~\ref{app:synthetic}). The resulting reward direction $v_u$ enables us to evaluate the estimator's ability to recover a ground-truth signal.

\section{Experiments}

\subsection{Conditions}
\label{sec:conditions}

We compare two estimators fitted on identical data with identical splits and seeds:

\begin{itemize}
\item \textbf{LoRe} --- the implementation released with
\citet{bose2025lorepersonalizingllmslowrank}: simplex-constrained weights,
cosine-to-$v_{\mathrm{sft}}$ anchor, column dropping, alternating optimization, run at
$\alpha = 0$, at which the objective reduces to the one published in the paper and which is
also the setting most favorable to it, since larger $\alpha$ drives the basis toward the
pretrained reward head\footnote{\citet{bose2025lorepersonalizingllmslowrank} report higher
accuracies for LoRe than we obtain here. Private correspondence with the authors indicates
that those numbers were affected by a bug in the released preprocessing; what we report is
the closest replication we were able to obtain. Both estimators are fitted on identical
corrected data with identical splits and seeds, so neither is advantaged.
Appendix~\ref{app:reimpl} gives the details.}
\item \textbf{RBD} --- the estimator of Section~\ref{sec:method}
\end{itemize}

We run both on Community Alignment with ranks $K \in \{1, 5, 8, 10, 20\}$ over five seeds.
% Each seed redraws both the train/validation/test split and the parameter initialization, so
% the reported spread covers the two things that vary between runs. 
On the synthetic control
of Section~\ref{sec:synthetic} we run on ranks $K \in \{1, 2, 3, 4, 8\}$. This range is
chosen to resolve the low-rank regime in which the two estimators differ.
% ; seed counts are
% stated per table. 
Two further conditions isolate where any gain originates:

\begin{itemize}
\item \textbf{Shared direction only.} Every $\delta_u$ is set to zero after fitting,
leaving the population direction $V\bar{w}$. Comparing this against the full model measures the per-user terms contribution.
\item \textbf{Few-shot adaptation.} Users are partitioned into seen and unseen. The model
is fitted on seen users; $V$ and $\bar{w}$ are then frozen and only $\delta_u$ is learned
for each unseen user, from budgets of 5, 10, 25 and 50 pairs. This condition uses 60 unseen
users over three seeds.
\end{itemize}

\subsection{Baselines}
\label{sec:baselines}

We compare against four reference directions, scored by
held-out pairwise accuracy:
% (defined in Section~\ref{sec:measures}).

\begin{enumerate}

\item\textbf{Base reward model} The base reward model is the pretrained Skywork reward head described in
Section~\ref{sec:method}. 
% Every accuracy we report
% is differenced against that seed's own base\,RM. 
% Its accuracy varies slightly across seeds because each seed draws a different test split; reporting each seed's difference from its own baseline removes that variation, where comparing absolute accuracies against one pooled
% baseline would not.
\item\textbf{LoRe}
LoRe is the baseline on real datasets, configured as in
Section~\ref{sec:conditions}. It fits the same reward family from the same embeddings under
the same splits and seeds as RBD.
% , so any difference is attributable to the
% decomposition.

\item\textbf{Global and other-user directions.} \textbf{Global} scores a user's held-out pairs
with one direction fitted across the whole population; \textbf{other-user} scores them with a direction fitted to a different, randomly chosen user; and \textbf{personal} scores them with a direction fitted to that user. The difference between \emph{personal} and \emph{other-user} gives the accuracy gain from reward model personalization.
% is the
% diagnostic of interest: if per-user structure exists and is recoverable, a user's own
% direction should outperform a stranger's on that user's pairs.

\item\textbf{Oracle} On the synthetic control, each user's true reward direction
is known. The oracle scores each user's pairs with the planted direction $v_u$ of
Equation~\ref{eq:planted}, upper-bounding the estimator.

\end{enumerate}

\subsection{Measures}
\label{sec:measures}

\textbf{Held-out pairwise accuracy} measures the fraction of held-out pairs
for which a direction $v$ satisfies $\Delta^\top v > 0$, where $\Delta$ is the difference
between the preferred and dispreferred embeddings, reported relative to base\,RM, (Section~\ref{sec:baselines}). We report the \textbf{delta contribution} $\delta_u$, measured as the difference between the full model's accuracy and the
shared-direction-only accuracy. $\delta_u=0$ means the per-user terms carry no signal that survives
to held-out data.

We additionally report \textbf{user-to-axis match}, the fraction of users whose recovered
reward direction $V(\bar{w} + \delta_u)$ has higher cosine with their own planted direction
than with any other user's. This measures identification rather than prediction, and unlike
accuracy it is invariant to the reparameterization of Section~\ref{sec:identifiability},
since it is computed on the induced per-user directions rather than on the columns of $V$.
%Chance is $1/N$ for $N$ users.

\subsection{Experimental Setup}
\label{sec:setup}

\paragraph{Data.} Community Alignment~\cite{zhang2026community} is filtered by English and users with at least 50 usable preference pairs. We then sub-sample to 200 users to limit the embedding cost (median 225 pairs/user).
PRISM~\cite{kirk2024prism} uses its own published train/test split with no
minimum-pairs filter, retaining all $1{,}029$ seen users and giving $15{,}754$
held-out pairs at a median of $15$ per user (minimum $8$, maximum $32$). PRISM
therefore has an order of magnitude less data per user, which bears on whether a
null there reflects the data or the sample size. The synthetic control of
Section~\ref{sec:synthetic} comprises 60 personas over 10 concept axes,
1{,}200 responses and 14{,}400 preference pairs.

\paragraph{Embeddings and splits.} We fit all models on frozen last-token hidden
states from Skywork-Reward-Llama-3.1-8B-v0.2~\cite{liu2024skywork}. 
% shared
% across all conditions since the backbone is frozen. 
We construct held-out sets
at the level of conversation turns rather than individual pairs, holding out
20\% of turns for validation and 20\% for test. 
% Each turn in Community Alignment
% yields three pairs sharing a prompt and preferred response, so splitting by pair
% would place siblings on both sides, letting a trained model memorize which
% response wins for a given prompt and score its sibling for free while an
% untrained baseline gains nothing. 
We select the learning rate and both ridge
strengths on that validation split at fixed $K = 8$ and reuse them at every rank
(Section~\ref{sec:method}). Appendix~\ref{app:synthetic} gives the splitting and refitting
procedure for the synthetic control.

\paragraph{Interpretation.} We analyse the population direction $V\bar{w}$ with
three techniques, all operating on $d$-dimensional directions: cosine alignment against a library of eleven named concepts, scored
against a null of $10{,}000$ random directions; the same scoring repeated on the
residual after removing $V\bar{w}$'s component along the pretrained head, which
separates alignment $V\bar{w}$ inherits from the head from alignment distinct to
it; and projection onto the decoder dictionary of a TopK sparse autoencoder
($16{,}384$ features, $k = 256$), compared by Jaccard overlap of the top-50
features. We interpret per-user directions only on the synthetic control, where
planted directions supply a ground truth. Appendix~\ref{app:controls} gives implementation details.

\section{Results}
\subsection{Interpreting the shared direction}
 \paragraph{RBD clears the base reward model; LoRe does not}
  \label{sec:ca-headline}

  RBD sits $+0.030 \pm 0.004$ above base\,RM at every rank, while LoRe sits
  $0.032 \pm 0.005$ below it. The gap of $+0.062$ is roughly fifteen times its own
  seed-to-seed spread. The difference is regularization rather than capacity. LoRe trains to $0.995$ accuracy
  and tests at $0.61$; RBD trains to $0.725$ and tests at $0.67$. Rank~1 makes this
  explicit: with a single basis direction and little room to memorize, LoRe nearly ties
  base\,RM ($-0.004$), and only falls away as rank increases. A basis that memorizes its training pairs gives back directions that describe those pairs, not what users prefer, and there is nothing in them to interpret.

  \begin{figure}[htb]
  \centering
  \includegraphics[width=0.9\linewidth]{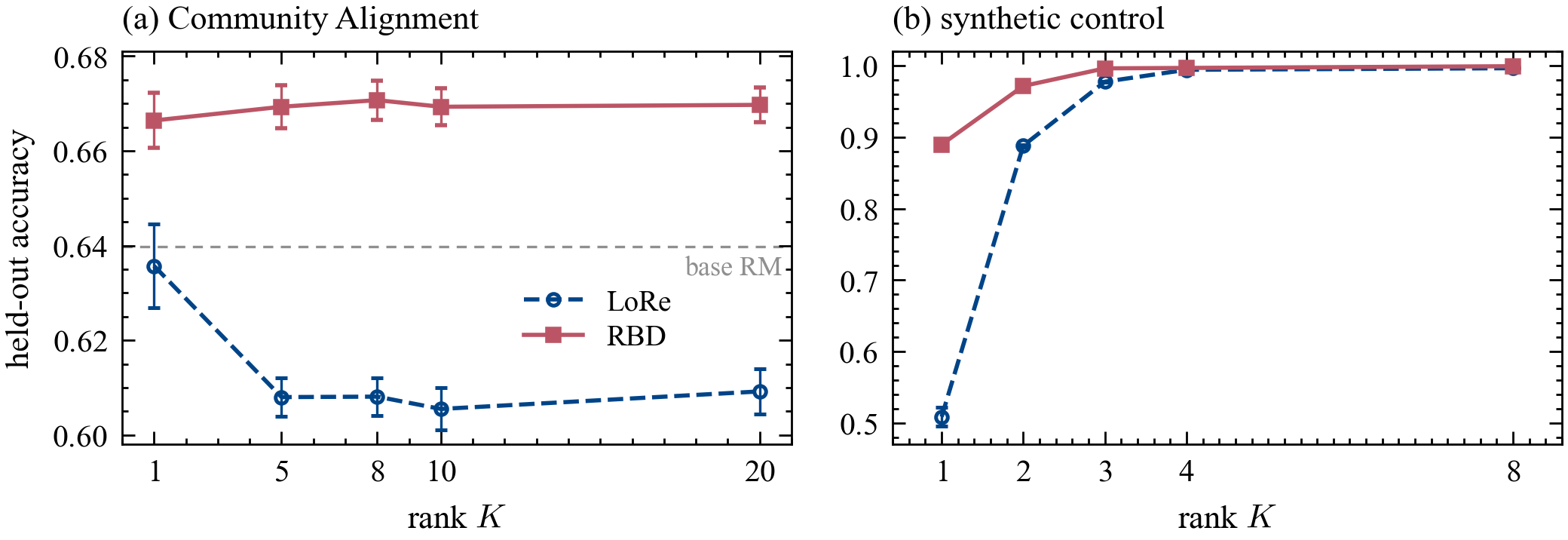}
  \caption{Held-out accuracy against rank, error bars one standard deviation over seeds.
  (a) Community Alignment, five seeds, with the base reward model at $0.6397$. Pooled
  over $K \geq 5$, RBD sits $+0.0301 \pm 0.0040$ above it and LoRe $-0.0320 \pm 0.0046$
  below, a gap of $+0.062$ that is roughly fifteen times its own seed-to-seed spread.
  (b) The synthetic control, three seeds. RBD at $K=1$ ($0.8897 \pm 0.0024$) already matches
  LoRe at $K=2$ ($0.8881 \pm 0.0015$), which is what signed weights predict, and accuracy
  saturates by $K=4$.}
  \vspace{-5pt}
  \label{fig:rank-panels}
\end{figure}

  \paragraph{The gain is entirely the shared direction}
  \label{sec:ca-ablation}
  Because RBD factors user weights as $\bar{w} + \delta_u$, setting every $\delta_u = 0$ isolates the population
  direction $V\bar{w}$. Doing so changes held-out accuracy by $-0.0007 \pm 0.0019$, pooled across ranks and seeds, which is indistinguishable from zero. At $K = 1$, every user's direction is a scalar
  multiple of the same vector, so a user whose total weight $\bar{w} + \delta_u$ is negative has its direction
  reversed and every pairwise decision inverted. This occurs for $13$ of $1,000$ user-fits across five seeds, and the measured $K=1$ delta contribution ($-0.0007 \pm 0.0007$) matches the pooled figure, so these flips are too rare to alter the conclusion. Two further experiments rule out that the ridge penalty $\lambda_{\delta}$ suppresses the per-user terms before they can contribute. Sweeping
  $\lambda_{\delta}$ over $\{0, 1, 2, 4, 6, 8, 10, 14, 16\}$ leaves the delta contribution
  within noise at every value, including $\lambda_{\delta} = 0$ with the penalty removed
  entirely ($-0.0012 \pm 0.0024$). Restricting to the most data-rich users, at $\geq 200$
  pairs ($n = 141$) and $\geq 250$ pairs ($n = 37$), likewise produces no personalization
  gain. 

  \paragraph{Adapting to unseen users does not improve accuracy}
  \label{sec:ca-unseen}

  Low-rank personalization helps fit a new user from a few preference pairs, and it is favorable to the per-user terms: $V$ and $\bar{w}$ are already fixed, so $\delta_u$ has only one user to explain and no shared structure to compete with. \iffalse If per-user structure is recoverable anywhere, it is here. \fi \iffalse If per-user structure is recoverable anywhere, it is here. \fi
\rev{This setting tests whether per-user adaptation improves
prediction within the learned shared basis.}
The $\bar{w} + \delta_u$ split tests the fit on 140 users, freezes $V$ and $\bar{w}$, and learns only a new $\delta_u$ for each of 60 held-out users (full results in Appendix~\ref{app:controls}, Table~\ref{tab:ca-unseen}). Adaptation never beats simply assigning a new user the population direction. It is worse at small-shot budgets (-0.0124 at 5 shots), where $\delta_u$ fits the noise, and converges to no difference as the ridge drives $\delta_u$ toward zero. The population direction
alone already beats base\,RM by $+0.029$ on users it never trained on, so the shared
direction works as well on users it never saw as those it was fitted on.

\subsection{Recovery of planted personalization signal}
\label{sec:synthetic-results}
\iffalse A null on real data cannot distinguish the absence of per-user structure from the estimator's failure to find it. We resolve this with two controls in which
per-user structure is present by construction: a synthetic population and
PRISM's own response embeddings with an injected preference direction. \fi \rev{A null on real data cannot distinguish the absence of per-user
structure from the estimator's failure to find it. We assess the
estimator's ability to recover known signals using two positive
controls: a synthetic population and PRISM's own response embeddings
with an injected preference direction. These controls establish
recoverability of the planted signals under the tested conditions,
but do not resolve this ambiguity for all forms of individual
preference.}

\paragraph{Synthetic data.}
Prompts are sampled from Community Alignment, and a
single model generates all text. For $37\%$ of distinct response pairs, one user prefers $A$ to $B$
while the other prefers $B$ to $A$, so no single direction can account for both
labels. The control therefore requires personalization: on a single split (seed
42), base\,RM scores $0.4798$, a single global direction $0.5220$, a user's own
direction $0.9105$, and the oracle (planted direction) $\mathbf{1.0000}$. A
single shared direction is barely above chance while the oracle is exact, so any
shortfall is attributable to the estimator. The \emph{personal} $-$
\emph{other-user} advantage is $\mathbf{+0.3337}$ on this split and $+0.40$ when
averaged over four seeds, compared to approximately $0.00$ on PRISM and Community
Alignment.

\begin{comment}
\begin{table}[h]
\caption{The synthetic control requires personalization. A single shared direction performs
barely above chance, while the oracle is exact, so any shortfall is attributable to the
estimator. Values are for a single split (seed 42); the \emph{personal} advantage is
$+0.33$ on that split and $+0.40$ averaged over four seeds, against approximately $0.00$ on
PRISM and Community Alignment.}
\label{tab:synth-reference}
\centering
\small
\begin{tabular}{lc}
\toprule
Direction & Held-out accuracy \\
\midrule
base\,RM (pretrained head)      & $0.4798$ \\
Single global direction         & $0.5220$ \\
User's own direction            & $0.9105$ \\
Oracle (planted direction)      & $\mathbf{1.0000}$ \\
\midrule
\emph{personal} $-$ \emph{other-user} & $\mathbf{+0.3337}$ \\
\bottomrule
\end{tabular}
\end{table}
\end{comment}

%\paragraph{Signed weights at rank 1 match the simplex at rank 2.}
Figure~\ref{fig:rank-panels}(b) reports held-out accuracy for both estimators across rank. At
$K=1$, LoRe is at chance ($0.5088$). With a single non-negatively weighted column,
every user's reward is a
positive multiple of the same vector, so a user who dislikes what another likes cannot be
represented. RBD reaches $0.8897$ on that same column. LoRe at $K=2$ scores $0.8881 \pm
0.0015$ against RBD at $K=1$ scoring $0.8897 \pm 0.0024$; the difference of $0.0016$ is
smaller than either standard deviation. This is what the parameterization predicts, since
signed weights on one column span the same set of reward functions as non-negative weights
on the two columns $\{v, -v\}$,  both reach exactly $\{cv : c \in \mathbb{R}\}$.

%\paragraph{Accuracy saturates at low rank.}
%Accuracy is saturated by $K=4$, with almost the entire climb occurring between $K=1$ and $K=2$, and the gap between estimators closing to $+0.003$. Above that rank the dataset is near-ceiling by construction and accuracy does not differentiate the two estimators.

\begin{figure}[htb]
  \centering
  \begin{minipage}{0.58\linewidth}
    \includegraphics[width=\linewidth]{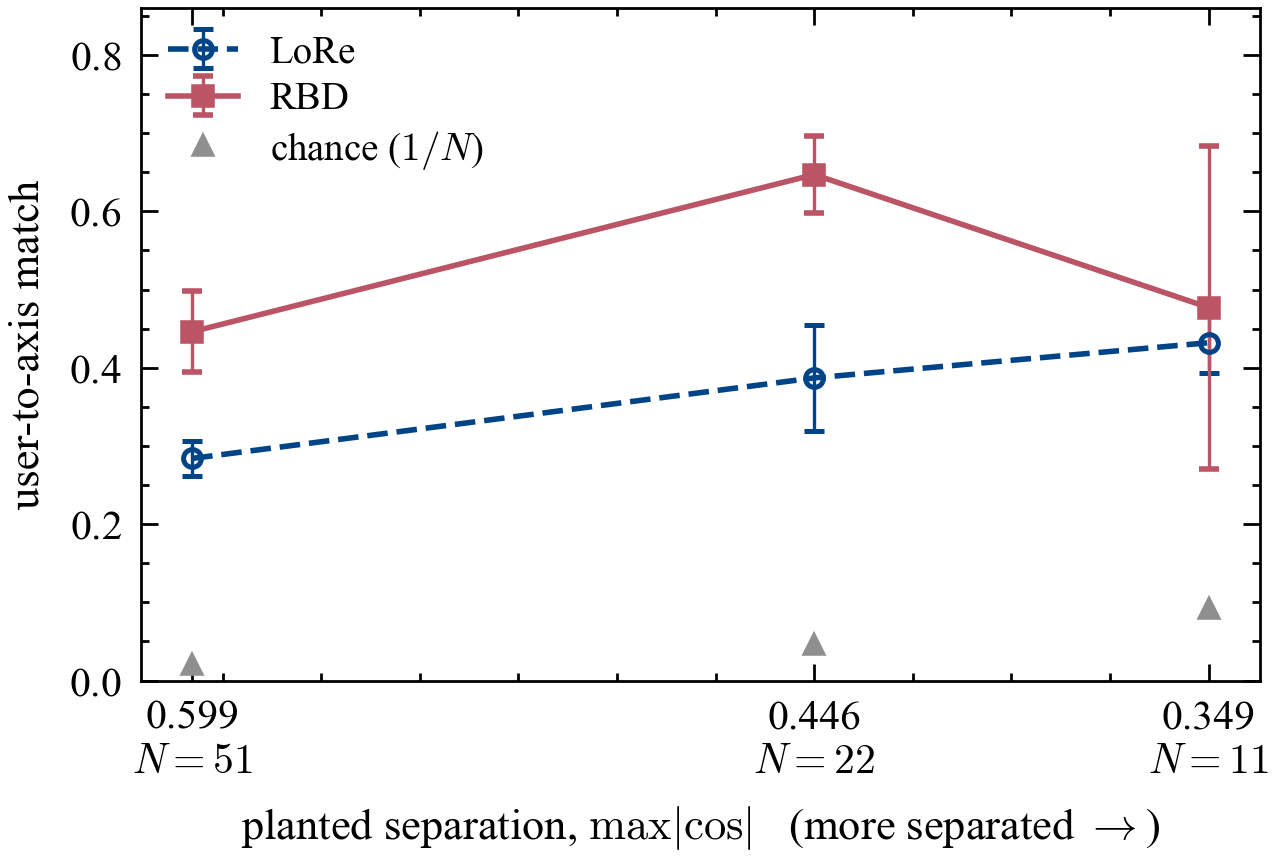}
  \end{minipage}\hfill
  \begin{minipage}{0.40\linewidth}
   \caption{User-to-axis match against planted separation, four seeds per point,
error bars one standard deviation. The chance is $1/N$ and rises as the user count
falls. Pooled over the 12
paired runs, RBD leads by $+0.156 \pm 0.048$ (standard error), positive in $10$
of $12$.}
    \label{fig:identification}
  \end{minipage}
\end{figure}

%\paragraph{Identification is a property of the estimator, not only of the data.}
User-to-axis match confounds how well an estimator recovers each user with the distinguishability of the planted users.
Figure~\ref{fig:identification} removes the confound by holding the planted separation fixed and varying only the estimator, paired on seed. LoRe improves monotonically as separation increases, while RBD peaks at intermediate
separation, though the spread at 11 users is too wide to determine whether
that peak reflects a real decline. If the drop is real, the shared basis there is
estimated from too little data, and RBD's advantage lies in modeling per-user
deviation from that basis. In absolute terms neither estimator reliably identifies users: even RBD's pooled match rate, well above chance, still assigns roughly
half of users to another user's direction.

\paragraph{Real PRISM data}
\label{sec:perturbation}
We test whether RBD recovers a personalization signal that we plant on PRISM's
real response-pair embeddings. Let $r \in \mathbb{R}^d$ be a unit-norm target
direction, either a random draw or a named concept from
Section~\ref{sec:population} (confidence, or fluency). For each real (chosen,
rejected) embedding difference, we strip the existing preference along $r$ and
inject a controlled signal ({signal\_scale} $\times$ median real diff norm),
under a bipolar ground truth of two user groups with opposing target directions
$\pm r$. At signal\_scale $=0$ (no injection), tested directions do not yield test accuracy above
chance ($0.539$ for a random direction). At signal\_scale $=0.5$
the test accuracy increases to $\sim0.83$, and the group-direction is $\sim0.99$, personal accuracy ($0.88$) exceeds global ($0.57$). The concept signals show the same trend (Table~\ref{tab:perturbation-bipolar},
Appendix~\ref{app:controls}). \iffalse The estimator cleanly recovers planted personalization, and finds no personalization on real data across all axes tested. \fi \rev{The estimator recovers the injected personalization signals
at the tested nonzero strengths, while our experiments on
unmodified real data show no reliable predictive benefit from
the fitted per-user terms.}

\subsection{Characterization of the population direction}

\paragraph{Alignment of the population direction with human-nameable concepts}
\label{sec:population}
 
We test whether $\bar w$ has interpretable content by testing its alignment with the base reward model (base\,RM) of Section~\ref{sec:baselines} and with an 11-concept library generated by contrastive prompting of Claude Sonnet 4.5, both scored against a random-direction null (941 users: the full eligible set at $\geq 100$ pairs, unsubsampled; 10,000 iterations, 5 seeds, $K=8$). $\bar w$ clears the null on 7 of 11 concepts. The residual is the component of $\bar w$ orthogonal to base\_head, i.e.\ what $\bar w$ adds beyond replicating the pretrained head. The residual clears only 5 of 7 concepts: confidence, helpfulness, fluency, diversity, and formatting. %Repetition and factuality drop out, so $\bar w$'s  alignment is inherited from its small overlap with base\_head rather than a distinct learned signal. %Fluency is the exception: its residual ($+0.063$) exceeds $\bar w$'s own raw cosine and runs opposite in sign to base\_head's fluency alignment ($-0.042$) i.e. the concept most clearly novel to $\bar w$. This pattern is stable across ranks at $K=1$ and $K=20$. We also test if length alone predicts preference. 
This pattern is stable across ranks at $K=1$ and $K=20$, though not across the generator that wrote the contrastive text: re-scoring the same direction against concept vectors from a second generator (Section~\ref{sec:generator-dependence}) preserves the ordering but moves \texttt{diversity} below the null, so individual significance calls should be read as generator-conditional.
The word-count accuracy on 66,540 held-out pairs is $0.5083$ ($p=0.008$), character-count accuracy is $0.4958$ ($p=0.22$), so length does not predict preference.

\begin{figure}[htb]
  \centering
 \includegraphics[width=0.9\linewidth]{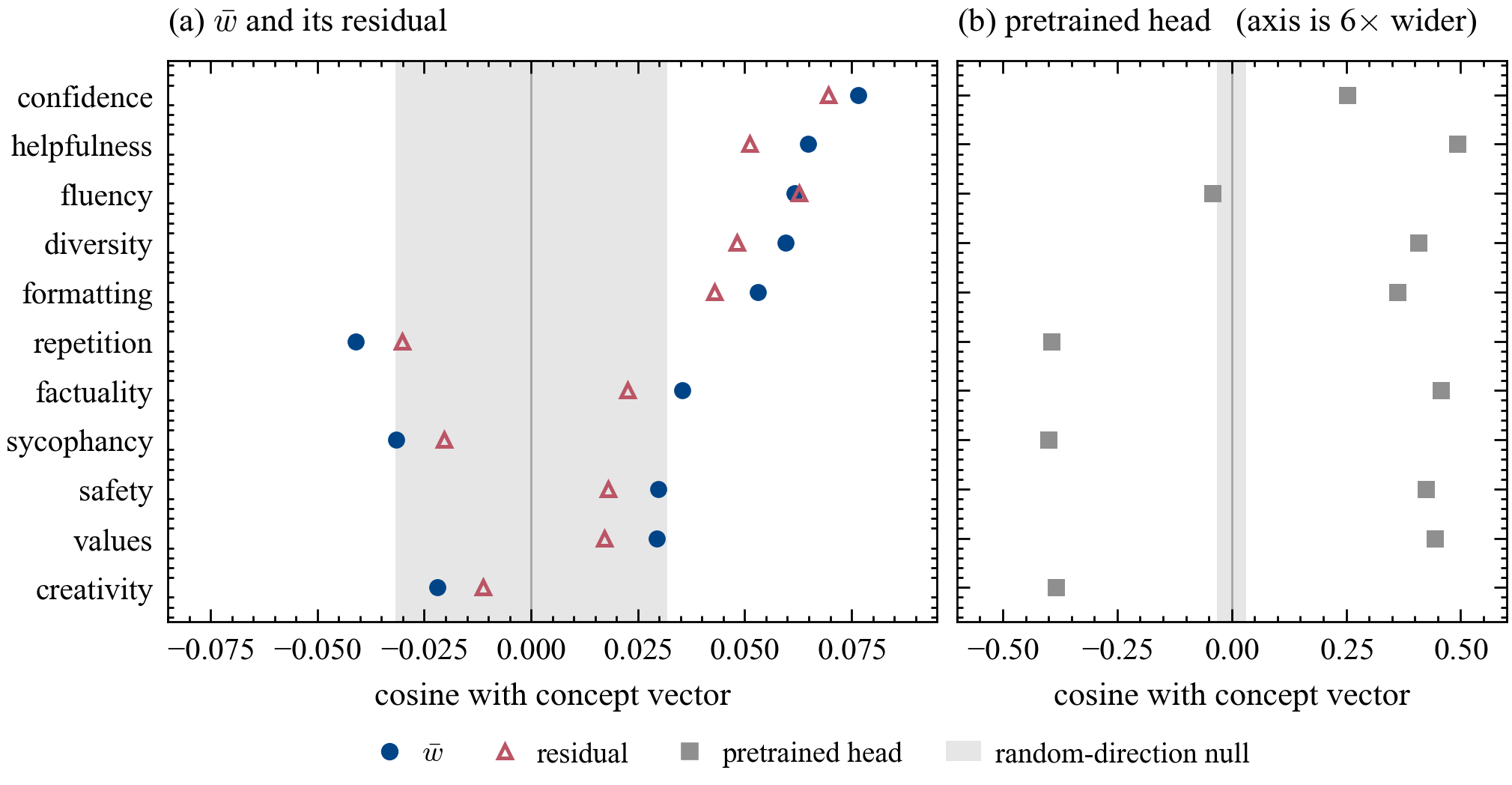}
  \caption{Concept alignment of the population direction, with the random-direction null
  band shaded; points outside the band clear the null at $p<0.05$. (a) $\bar w$ and its
  residual (b) base\_head on its own scale, an
  order of magnitude larger. $\bar w$ clears the null on 7 of 11 concepts and the residual
  on 5; fluency is the only concept whose residual ($+0.063$) exceeds $\bar w$'s own cosine
  and runs opposite in sign to base\_head. Full numbers in
  Table~\ref{tab:concept-alignment-full}.}
  \label{fig:concepts}
\end{figure}

\paragraph{The population direction is not the pretrained head}
\label{sec:sae}
We test whether the population direction $V\bar{w}$ is a distinct learned axis or
the pretrained head under a change of basis, running six experiments on PRISM and
Community Alignment at the tuned configuration of Section~\ref{sec:method}
($K=8$, seed 42, $\lambda_{\mathrm{pop}}=0.01$, $\lambda_\delta=10$, setup in
Appendix~\ref{app:controls}).

\begin{table}[htb]
\centering
\small
\setlength{\tabcolsep}{4pt}
\begin{minipage}{0.46\linewidth}
\centering
\begin{tabular}{lcc}
\toprule
 & PRISM & CA \\
\midrule
$\cos(V\bar{w},\texttt{base\_head})$ & 0.0118 & 0.0548 \\
score corr.\ vs.\ \texttt{base\_head} & 0.630 & 0.749 \\
\midrule
accuracy, \texttt{base\_head} & 0.5902 & 0.6364 \\
accuracy, $V\bar{w}$ & 0.6327 & 0.7050 \\
accuracy, residual & 0.6293 & 0.6633 \\
\midrule
Jaccard, $V\bar{w}$ vs.\ head & 0.010 & 0.020 \\
Jaccard, residual vs.\ head & 0.010 & 0.020 \\
Jaccard, residual vs.\ $V\bar{w}$ & 1.000 & 0.754 \\
\bottomrule
\end{tabular}
\end{minipage}\hfill
\begin{minipage}{0.50\linewidth}
\caption{The population direction against the pretrained head, on PRISM and
Community Alignment (CA). Accuracy is absolute held-out pairwise accuracy on a
single seed-42 split, from a fit made separately from those of Figure~\ref{fig:rank-panels}. Jaccard is the top-50 overlap in the sparse
autoencoder dictionary.}
\label{tab:sae}
\end{minipage}
\end{table}

% Table~\ref{tab:sae} gives one answer three ways. 
$V\bar{w}$ and
\texttt{base\_head} share almost no top-50 dictionary features, and
removing the head component changes neither the feature set nor the accuracy, so the gain is not coming from the head. $V\bar{w}$ predicts held-out
pairs better than the head on both datasets while nearly orthogonal to it as a
vector. The score correlations show the two still rank pairs similarly, so
the accuracy gap reflects a cleaner separation. Three further results in
Appendix~\ref{app:controls} support this. No single fixed direction reaches
$V\bar{w}$. Sign-flipping the pair labels and refitting yields a direction
uncorrelated with the real $V\bar{w}$ at chance accuracy, and a known non-head
direction injected into the data is recovered off the head. Transfer across
datasets is partial, with the PRISM-fit direction scoring below the head on
Community Alignment.
 
\paragraph{Concept vectors depend on the generating model}
\label{sec:generator-dependence}
Concept vectors are in part properties of the model that wrote the contrastive text. We generated six concepts twice from byte-identical prompts, varying only the generating model, and compared directions against a within-generator noise floor by splitting each concept's pairs in half
and estimating the direction from each half independently.

\begin{table}[htb]
\centering
\small
\setlength{\tabcolsep}{4pt}
\begin{minipage}{0.45\linewidth}
\centering
\begin{tabular}{lcc}
\toprule
Concept & Cross-gen.\ $\cos$ & Split-half \\
\midrule
\texttt{repetition} & $\mathbf{-0.248}$ & $0.850 \,/\, 0.945$ \\
\texttt{confidence} & $+0.758$ & $0.972 \,/\, 0.952$ \\
\texttt{fluency}    & $+0.775$ & $0.971 \,/\, 0.942$ \\
\texttt{diversity}  & $+0.831$ & $0.956 \,/\, 0.986$ \\
\texttt{formatting} & $+0.930$ & $0.990 \,/\, 0.981$ \\
\texttt{creativity} & $+0.953$ & $0.961 \,/\, 0.981$ \\
\midrule
Mean & $+0.667$ & $\approx 0.95$ \\
\bottomrule
\end{tabular}
\end{minipage}\hfill
\begin{minipage}{0.51\linewidth}
\caption{The same concept yields different directions for identical text 
depending on the generating model. Split-half reliability is $0.85$--$0.99$ throughout, so the cross-generator
disagreement is not estimation noise.}
\label{tab:generator}
\end{minipage}
\end{table}

\begin{figure}[htb]
  \centering
 \includegraphics[width=0.7\linewidth]{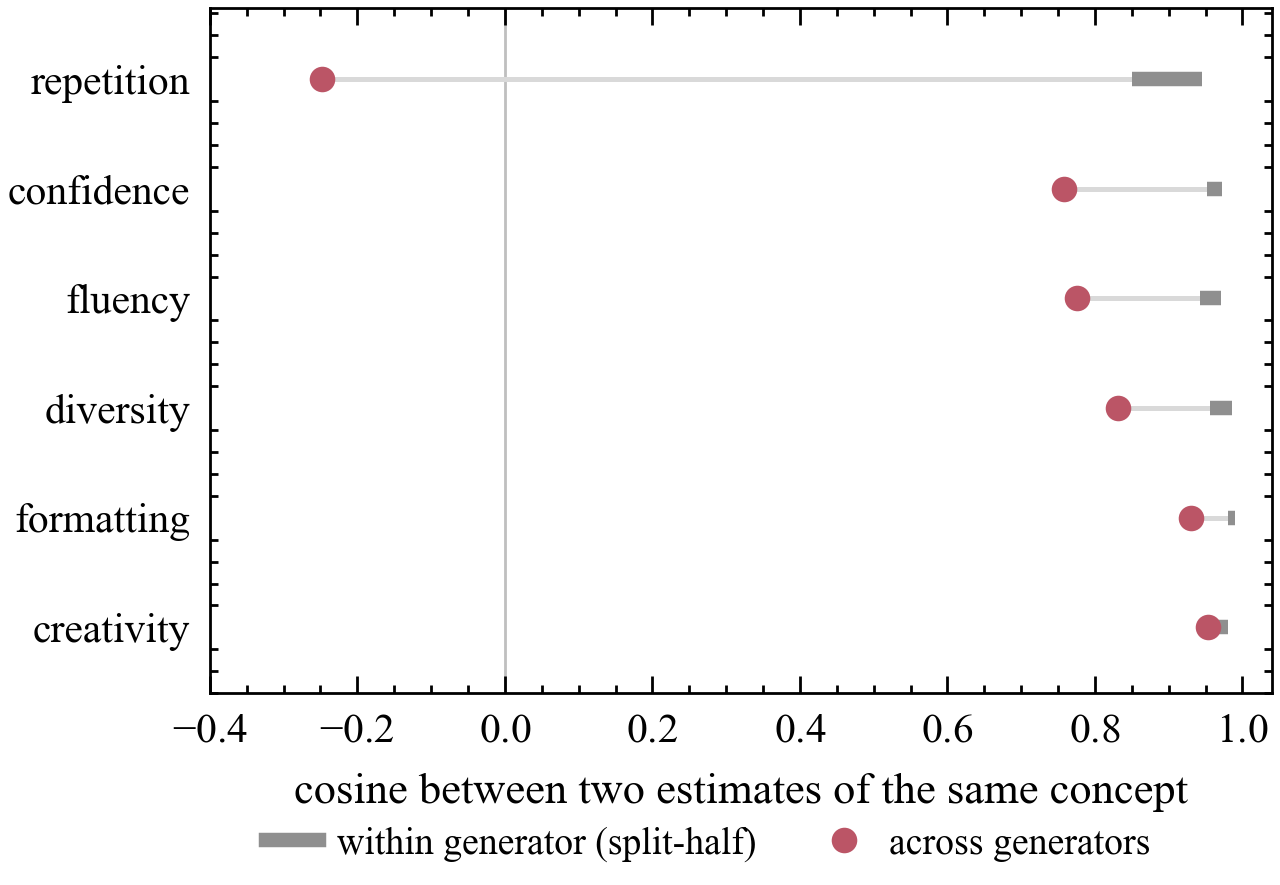}
  \caption{Cross-generator disagreement for concept vectors}
  \label{fig:concepts_cross_gen}
\end{figure}

For \texttt{repetition}, the two generators produce opposite-signed directions,
with both sets of contrastive text correct on inspection as seen in Figure~\ref{fig:concepts_cross_gen}. This bears on the
concept alignments in Figure~\ref{fig:concepts}, whose magnitudes of
$0.03$--$0.08$ are below the mean cosine of $0.667$, at which the concept axes
measuring them reproduce across generators. The per-concept alignments of this size are best read
as generator-conditional rather than as properties of the concept.
For \texttt{repetition}, its negative cross-generator
cosine means the sign of its reported alignment is not stable under a change of
generator. Concept-alignment results of this kind should therefore either fix and
report the generating model, or average the concept direction over several
generators before measuring alignment against it.

\paragraph{Stability of the learned direction}
\label{sec:stability}
We test the reproducibility of $V\bar w$ and the per-user directions $V(\bar w + \delta_u)$ using the rotation-invariant unit directions $v_{\text{pop}} =
\text{unit}(V\bar w)$ and $v_u = \text{unit}(V(\bar w + \delta_u))$
(Section~\ref{sec:identifiability}). At convergence, we see that $v_{\text{pop}}$ is stable
across random seeds (cosine $0.9999$ over 10 seeds at a fixed split) and across
train/test splits (cosine $0.8649$ over 5 splits at a fixed initialization), both
above a random-direction baseline. Per-user directions is similar to 
$v_{\text{pop}}$, so $\delta_u$ carries no reproducible user-specific signal
(see results in Appendix~\ref{app:stability}).
 
%Through the experiments and results shown above, we have introduced RBD, our new mechanistic interpretability technique for reward bases. RBD decomposes a multi-user reward model into a shared population direction and a set of per-user directions, thereby allowing interpretation of both. Applying our method to two real preference datasets, PRISM and Community Alignment, reveals two findings.

%The first is that across both datasets, the per-user directions carry little to no signal that survives to held-out data. We show that zeroing every per-user direction results in a negligible drop in accuracy, and few-shot adaptation to unseen users never beats assigning them the population direction. To ensure that this null result was not a shortcoming of our method, we tested our method on a synthetic population with planted per-user directions, where RBD recovers a substantial personal-direction advantage. Our experiments with synthetic data demonstrate that RBD finds per-user structure where it is known to exist. The fact that we find no significant per-user direction in PRISM or Community Alignment, points to the fact that the null result is a property of the underlying data rather than a failure to recover it.

\section{Discussion}
We introduced Reward Basis Decomposition (RBD), which enables independent
analysis of the shared population direction and the individual per-user
directions. \iffalse We validated RBD on a synthetic dataset, verifying that it recovers
artificially injected individual-user directions, and found that neither PRISM
nor Community Alignment had individual-user structure to discover when
decomposed this way. \fi \rev{We validated RBD on a synthetic dataset, verifying that it
recovers artificially planted per-user directions. On PRISM and
Community Alignment, the fitted per-user terms did not reliably
improve held-out prediction beyond the shared population direction
under our setup.} The population direction predicts held-out pairs more
accurately than the head on both datasets and shares almost none of its top
sparse-autoencoder features. The concept library we test partially
explains it, but the residual analysis shows most of that alignment is inherited
from its small overlap with the head.

\paragraph{Broader Impact.} Our method introduces a new object for future
mech-interp work to study, enabling researchers and model providers to better
understand what a shared reward model has learned and allowing practitioners to
gauge whether a preference dataset supports individual-level reward modeling,
which can catch bad personalization claims before deployment. A malicious actor
could also use it to poison a dataset until it appears to encode individual
preferences, yielding a personalization system that does not serve individual
users faithfully.
\iffalse
\paragraph{Limitations.} RBD recovers per-user preferences only in datasets
constructed as a best-case scenario with planted directions. On PRISM and
Community Alignment the per-user contribution is minuscule and few-shot
adaptation to unseen users never outperforms the shared direction alone, and
even in the favorable synthetic setting RBD matches only about half of users to
their own direction. \fi

\paragraph{Limitations.}
\rev{The synthetic controls establish recoverability of the planted
signals under the tested conditions, but do not rule out weaker
or differently encoded individual preferences in real data.
Our null results therefore concern what RBD can recover with
the chosen representation, model class, and available data,
rather than the absence of individual preferences. On PRISM
and Community Alignment, the fitted per-user contribution is
minuscule and few-shot adaptation to unseen users never
outperforms the shared direction alone. Even in the favorable
synthetic setting, RBD matches only about half of users to
their own direction.}

\paragraph{Future Work.} Our experiments are limited to PRISM, Community
Alignment, and the Skywork-Reward-Llama-3.1-8B backbone, and could be
substantiated with more diverse datasets and reward backbones. One future
direction is to modify the LoRe loss objective to encourage sparsity, and by
extension monosemanticity, of learned features, so that they map more tractably
to human-interpretable concept labels and are less sensitive to the choice of
concept-generating model.

\begin{ack}
This research was conducted as part of the PRISM AI Safety Research Fellowship,
a program of Women Who Do Data (W2D2), with support from the Long-Term Future
Fund (LTFF). We thank Hassan Suliman for help with the sparse autoencoder
experiments and with formatting the paper, Archana Vaidheeswaran for reviewing
multiple drafts of the paper, Soham Chatterjee for feedback during the
mid-program updates, Ayesha Imran for coordinating the work, and ChengCheng Tan
for her help in polishing the paper.
\end{ack}

\bibliographystyle{plainnat}  
\bibliography{refs}

%%%%%%%%%%%%%%%%%%%%%%%%%%%%%%%%%%%%%%%%%%%%%%%%%%%%%%%%%%%%

\appendix
\section{Implementation of Reward Basis Decomposition}
\subsection{Tuning protocol}
\label{app:tuning}

We sweep the learning rate over $\{10^{-4}, 10^{-3}, 10^{-2}, 10^{-1}\}$, then grid both
penalties over $\{0.001, 0.01, 0.1, 1, 10\}$ at fixed $K = 8$, then re-check the learning
rate at the selected penalties: 33 runs in total. The selected values are reused at every
rank, so regularization strength does not depend on $K$.

Run-to-run variation from initialization alone is roughly $\pm 0.03$ in validation accuracy,
an order of magnitude larger than the differences between grid cells ($\approx 0.001$).
Every configuration is therefore initialized identically before comparison; without this the
sweep selects on initialization noise.

\subsection{Identifiability}
\label{app:identifiability}

Section~\ref{sec:identifiability} states the invariance; this appendix provides the argument
and its implications for reporting.

Write the objective in terms of the products it actually depends on. Both the likelihood and
$\mathcal{R}$ of Equation~\ref{eq:ridge} involve $V$ and the weights only through $V\bar{w}$
and $V\delta_u$. For any invertible $R \in \mathbb{R}^{K \times K}$, substituting $V \mapsto
VR$ and $w \mapsto R^{-1}w$ leaves both products unchanged, since $(VR)(R^{-1}\bar{w}) =
V\bar{w}$, and therefore leaves the objective unchanged. The fit is thus determined only up
to $R$: the column space of $V$ and the induced reward functions are identified, while the
individual columns of $V$ are not.

The consequence is that per-column quantities computed on the raw $V$ carry no meaning for
RBD. A basis column's cosine with a concept vector, for instance, can be given any value by
a suitable choice of $R$ without changing a single prediction. We therefore report only
transform-invariant quantities: the shared direction $V\bar{w}$ and the per-user directions
$V(\bar{w} + \delta_u)$. Where interpretable axes are required, we take the principal axes
of the centred per-user reward deviations $\{V\delta_u\}$ by SVD, followed by a varimax
rotation; these are determined by the deviations themselves rather than by the arbitrary
basis in which they are expressed.

\subsection{Construction of the synthetic control}
\label{app:synthetic}

\paragraph{Concept library.} Reward directions are defined by contrastive generation: for a
named concept, a language model writes response pairs that differ along that concept alone,
and the concept vector is the difference in mean embedding between the high and low poles. All contrastive text for the synthetic control, and the persona response pool itself, is generated by Claude Opus 4.6; the eleven-concept library used for the alignment analysis of Section~\ref{sec:population} was generated earlier by Claude Sonnet 4.5. Because concept directions turn out to depend on the generating model, a single generator is used throughout the construction rather than mixing the two.
Persona axes must be mutually distinct, or two personas built from different concepts are
the same user under two names. We measure the effective rank of a concept set as $\exp$ of
the entropy of its singular-value spectrum. On that measure, the eleven-concept library used
for the alignment analysis of Section~\ref{sec:population} spans an effective rank of only
$3.48$: the five quality concepts have effective rank $1.54$ between them, with
\texttt{safety} and \texttt{values} at $\cos = 0.98$. We therefore drop the quality cluster,
retain \texttt{diversity} as its representative, and add six stylistic concepts, giving
twelve concepts at effective rank $5.19$. Personas are built over the ten most independent
of these.

\paragraph{Personas.} Each synthetic user $u$ receives a sparse signed weight vector
$\omega_u$ over the $m$ concept axes, with a small set of preferred concepts taking positive
weight and a disjoint set of dispreferred concepts taking negative weight. Candidates are
accepted by rejection sampling subject to a maximum pairwise cosine between the reward
directions they induce. That test must be applied to the direction the labels are actually
generated from, given in Equation~\ref{eq:planted} below, rather than to $\omega_u$ or to
the unit concept basis. The per-concept scales differ by a factor of five across this pool,
and enforcing separation on the unscaled basis admits persona pairs at $\cos \approx 0.95$
in the space the labels occupy.

\paragraph{Shared response pool.} Generating each persona its own responses would not test
personalization. No two users would ever judge the same item, so ``user $u$ prefers the
response written for $u$'' is separable by a single global direction that merely detects
which persona a response was written for. Instead each prompt receives one pool of responses
spanning every concept pole, high and low, and every user judges that same pool. The cost of
the pool is then independent of the number of users. With $60$ prompts and $10$ concepts
this gives $60 \times 20 = 1{,}200$ responses.

\paragraph{Labels.} Let $C \in \mathbb{R}^{d \times m}$ hold the unit concept vectors and
let $z(y) \in \mathbb{R}^m$ be the projection of a response embedding onto $C$, standardized
across the pool so that concepts contribute comparably. User $u$'s utility is
$\mathcal{U}_u(y) = \omega_u^\top z(y)$, so the induced reward direction is
\begin{equation}
v_u \;=\; C\,(\omega_u \oslash s),
\label{eq:planted}
\end{equation}
where $s$ is the vector of per-concept standard deviations and $\oslash$ is elementwise
division. Preference pairs are drawn within a prompt and retained only when the utility gap
exceeds a margin, so that pairs the persona is indifferent about are excluded rather than
labelled by numerical noise. Because $v_u$ is known in closed form, it serves as an oracle:
any shortfall against it is attributable to the estimator and not to the construction.

\paragraph{Splits.} Each user draws many pairs from the same set of prompts. Splitting by
pair would place the same response texts on both sides of the split, letting a trained model
memorize which response wins for a given prompt while an untrained baseline gains nothing.
Held-out sets are therefore constructed by whole prompt: test prompts, and every response
written for them, are unseen at training time. The identification results additionally vary the planted separation between
personas, refitting the persona set and the preference pairs at each level while
holding the response pool fixed, and the pool is generated per concept rather
than per persona, so it does not depend on the population.
\section{The LoRe baseline}
\label{app:reimpl}

LoRe results in this paper come from the public repository released with
\citet{bose2025lorepersonalizingllmslowrank} to reproduce that paper's experiments, used
directly rather than reimplemented, so the baseline is the published method itself rather
than our rendering of it. The comparison requires both estimators to be fitted on identical
data, splits and seeds, and two corrections to that code were required before this was
possible. Neither touches the method; both change the baseline's numbers, which are
consequently not directly comparable to those reported in the original paper. Private
correspondence with the authors indicates that the discrepancy in reported accuracy is
attributable to the preprocessing issue described below.

\paragraph{The reference direction was not the reward head.} The released implementation
regularizes the basis toward $v_{\mathrm{sft}}$, described as the pretrained reward model's
direction. In the released code this vector is extracted by loading the checkpoint with a
bare backbone class and taking the last linear layer reachable from the module tree. For a
sequence-classification checkpoint that class does not load the trained scoring head at all:
the head's weights are dropped as unexpected keys, and the last linear layer in the
remaining module tree is an internal feed-forward projection of the final transformer block.
A single column of that matrix was being used as the reward direction. It is not one.

We load the scoring head directly instead. This changes two things at once. The base reward
model, against which all accuracies are reported, is now the actual reward head rather than
an arbitrary internal column. And the anchor now pulls the basis toward the pretrained
reward direction, which is what the penalty was intended to do, so $\alpha$ has the meaning
ascribed to it. All results in this paper use the corrected extraction.

\paragraph{A formatting artifact in the preprocessing.} The released preprocessing
serialises the chosen response as a string but the rejected response as a list. The chat
template consequently renders the rejected side as a Python list literal, with brackets,
quotes and commas, against clean prose on the chosen side. Every preference pair is then
separable by a formatting fingerprint rather than by preference. This has two consequences,
which are the same effect seen from two directions: held-out accuracy is inflated, because
the fingerprint is perfectly predictive and present in the held-out split as well; and the
learned basis collapses, because every basis column converges on the single direction that
encodes the fingerprint.

\paragraph{The corrected procedure.} We format both sides identically as a single string. A
second artifact is handled at the same time: $151$ of $12{,}999$ turns have an empty
rejected list, which the original code embedded as the literal \texttt{[]}. Those turns
carry no valid rejected response and are discarded rather than embedded.

\paragraph{Effect.} Under the corrected procedure, held-out accuracy falls from approximately
$0.98$ to approximately $0.59$, at which point the estimator matches the base reward model
rather than exceeding it, and the basis no longer collapses. Both estimators are fitted on
the corrected data throughout.

\section{Additional results and controls for the population-direction
analysis}
\label{app:controls}

\paragraph{Interpretation details.} Every concept in the eleven-concept library
is obtained as a mean-difference vector over contrastively generated response
pairs. A word- and character-count control tests whether response length alone
accounts for any alignment found. One fit departs from the setup of
Section~\ref{sec:setup}: the concept alignment uses all $941$ Community
Alignment users with at least $100$ pairs rather than the $200$-user subsample.

\paragraph{Setup.} PRISM: 1{,}029 users, 15{,}754 held-out pairs, no
minimum-pairs filter; Community Alignment: the 200-user subsample and
turn-level split of Section~\ref{sec:setup}, 9{,}010 held-out pairs. The
\emph{residual} is $V\bar{w}$ with its component along \texttt{base\_head}
removed. Directions are compared by cosine, by held-out pairwise accuracy,
and by the Jaccard overlap of their top-50 features under the sparse
autoencoder of Section~\ref{sec:setup}. On PRISM the residual's scores still
correlate $0.553$ with the head's, against $0.078$ on Community Alignment, so
the separation there is weaker in score space than in vector or feature space.

\paragraph{Single fixed directions.} The mean-difference direction of
preferred minus not selected embeddings aligns with the head ($\cos$ 0.428
PRISM, 0.451 Community Alignment) and predicts at the head's level (0.5909,
0.6378); the best single logistic direction fit on train pairs reaches 0.5760
and 0.6494. Both sit below $V\bar{w}$ (0.6327, 0.7050). Removing the head
component from $V\bar{w}$ also leaves its norm essentially unchanged (0.9985
on Community Alignment).

\paragraph{Controls.} Sign-flipping each pair's label with probability 0.5
and refitting gives a direction uncorrelated with the real $V\bar{w}$
($\cos = 0.004$ on PRISM, $-0.011$ on Community Alignment), at chance
held-out accuracy (0.5056, 0.4984) with no concept clearing the
random-direction null of Section~\ref{sec:population}; under the tuned ridge
$\|\bar{w}\|$ barely moves (1.136 to 1.124 on Community Alignment), so the
null presents as chance prediction rather than a norm collapse. Injecting a
known non-head direction (fluency, $\cos$ to head $-0.041$) and refitting
recovers it at $\cos$ 0.9798 on PRISM and 0.9964 on Community Alignment at
the fit learning rate, and above 0.99999 at learning rate $10^{-2}$, off the
head in every case.

\paragraph{Cross-dataset transfer.} The PRISM and Community Alignment
population directions have cosine 0.033 and score correlation 0.610. On
Community Alignment held-out pairs the Community-Alignment-fit direction
scores 0.7050, the head 0.6364, and the PRISM-fit direction 0.5984. The
PRISM direction carried to Community Alignment keeps a correlated signal but loses its accuracy advantage.

\paragraph{Full few-shot adaptation table.} Table~\ref{tab:ca-unseen} shows the full
few-shot adaptation results from Section~\ref{sec:ca-unseen}.

\begin{table}[htb]
\caption{Few-shot adaptation for 60 unseen users, three seeds.}
\label{tab:ca-unseen}
\centering
\begin{tabular}{rcccc}
\toprule
shots & base\,RM & $\bar{w}$ only & adapted & adapted $-$ $\bar{w}$ \\
\midrule
5  & 0.6376 & 0.6664 & 0.6540 & $-0.0124 \pm 0.0129$ \\
10 & 0.6376 & 0.6664 & 0.6550 & $-0.0114 \pm 0.0097$ \\
25 & 0.6376 & 0.6664 & 0.6630 & $-0.0034 \pm 0.0036$ \\
50 & 0.6376 & 0.6664 & 0.6611 & $-0.0053 \pm 0.0034$ \\
\bottomrule
\end{tabular}
\end{table}

\paragraph{Recovery of planted personalization signal}
Table~\ref{tab:perturbation-bipolar} shows the results for signal recovery on real PRISM data in Section ~\ref{sec:synthetic-results}; all three sources display the same recovery pattern.

\begin{table}[htb]
\caption{Synthetic recovery under bipolar ground truth, at pairs\_per\_user$=15$, across signal sources and injection strengths.}
\label{tab:perturbation-bipolar}
\centering
\small
\begin{tabular}{llccccc}
\toprule
Source & Signal & Test acc. & Subspace align. & Group dir. match & Personal & Global \\
\midrule
Random     & 0.0 & 0.5392 & 0.0023 & 0.5600 & 0.5420 & 0.5729 \\
Random     & 0.5 & 0.8269 & 0.6709 & 0.9900 & 0.8766 & 0.5729 \\
Random     & 4.0 & 0.8262 & 0.0991 & 1.0000 & 0.8984 & 0.5315 \\
\addlinespace
Concept (confidence) & 0.0 & 0.5403 & 0.0020 & 0.4650 & 0.5389 & 0.5749 \\
Concept (confidence) & 0.5 & 0.8282 & 0.6704 & 0.9900 & 0.8858 & 0.5741 \\
Concept (confidence) & 4.0 & 0.8372 & 0.0771 & 0.9950 & 0.8984 & 0.5209 \\
\addlinespace
Concept (fluency) & 0.0 & 0.5413 & 0.0059 & 0.5250 & 0.5414 & 0.5724 \\
Concept (fluency) & 0.5 & 0.8274 & 0.6353 & 0.9800 & 0.8778 & 0.5724 \\
Concept (fluency) & 4.0 & 0.8353 & 0.0837 & 0.9950 & 0.8984 & 0.5308 \\
\bottomrule
\end{tabular}
\end{table}

\paragraph{Full concept-alignment table.} Table~\ref{tab:concept-alignment-full} shows all 11 concepts in Figure~\ref{fig:concepts} in
Section~\ref{sec:population}, including the four (sycophancy, safety, values, creativity)
that clear the null for base\_head but not the $\bar w$ or the residual.

\begin{table}[htb]
\centering
\caption{Concept alignment for $\bar w$ vs. base\_head, and the residual orthogonal to
base\_head. \textbf{Bold} marks cosines that clear the random-direction null at $p<0.05$.}
\label{tab:concept-alignment-full}
\begin{tabular}{lccc}
\toprule
Concept & $\bar w$ cos & base\_head cos & Residual cos \\
\midrule
confidence   & \textbf{+0.0766} & \textbf{+0.2523} & \textbf{+0.0696} \\
helpfulness  & \textbf{+0.0648} & \textbf{+0.4935} & \textbf{+0.0511} \\
fluency      & \textbf{+0.0616} & \textbf{$-$0.0416} & \textbf{+0.0628} \\
diversity    & \textbf{+0.0595} & \textbf{+0.4083} & \textbf{+0.0482} \\
formatting   & \textbf{+0.0530} & \textbf{+0.3621} & \textbf{+0.0430} \\
repetition   & \textbf{$-$0.0411} & \textbf{$-$0.3939} & $-$0.0302 \\
factuality   & \textbf{+0.0353} & \textbf{+0.4574} & 0.0226 \\
sycophancy   & $-$0.0315 & \textbf{$-$0.4007} & $-$0.0204 \\
safety       & 0.0297 & \textbf{+0.4235} & 0.0180 \\
values       & 0.0295 & \textbf{+0.4444} & 0.0172 \\
creativity   & $-$0.0219 & \textbf{$-$0.3831} & $-$0.0113 \\
\bottomrule
\end{tabular}
\end{table}

\paragraph{Seed and split stability of the learned direction.} 
\label{app:stability}
We compute the pairwise cosine similarity for
$v_{\text{pop}} = \text{unit}(V\bar w)$ across 10 seeds at a fixed split, and across 5 splits
at a fixed initialization. The baseline is 200 isotropic random unit directions in the
same 4096-dimensional space. At 1,500 iterations, training loss is still decreasing, appearing to be a flat optimum. On extending to 10,000 iterations, the seed spread collapses to 0.9999. At convergence, initialization does not meaningfully affect the learned direction. The split check gives cosine 0.8649, which is above the random baseline but below the seed-reproducibility ceiling. Per-user directions is similar to $v_{\mathrm{pop}}$ showing
$\delta_u$ carries no reproducible user-specific signal.

\begin{table}[htb]
\caption{Direction stability across seeds and splits (Community Alignment, 941 users: the
full eligible set at $\geq 100$ pairs, unsubsampled).}
\label{tab:stability}
\centering
\small
\setlength{\tabcolsep}{4pt}
\begin{tabular}{lccccc}
\toprule
Check & Iters & $v_{\mathrm{pop}}$ cosine (range) & Per-user cosine & Test acc. & $\cos(v_{\mathrm{pop}}, \mathrm{base\_rm})$ \\
\midrule
Seed  & 1{,}500  & 0.82 (0.68--0.97)         & --     & $0.6796\pm0.0006$ & -- \\
Seed  & 10{,}000 & 0.9999 (0.9997--1.0000)   & 0.9998 & $0.6814\pm0.0003$ & $0.0301\pm0.0001$ \\
Split & 10{,}000 & 0.8649 (0.8601--0.8712)   & 0.8649 & $0.6788\pm0.0019$ & $0.0276\pm0.0070$ \\
\bottomrule
\end{tabular}
\end{table}

A disjoint-halves check tests whether this reflects sampling noise: two independent, non-overlapping halves of the 941 eligible users (470/471 each) fit separately give raw $\cos(v_{\mathrm{pop},A}, v_{\mathrm{pop},B}) =
0.372\pm0.074$. Behaviorally, the two halves are almost identical: half $A$'s direction scores half $B$'s held-out pairs within $0.002$ of half $B$'s own accuracy ($0.6697$ vs.\ $0.6717$), and the two directions' scores correlate at $+0.924$ across held-out pairs. $v_{\text{pop}}$ is stable as a function with accuracy within $0.002$, score correlation $0.92$, but not as a vector, since the raw cosine between the two halves is only $0.372$. 

\begin{figure}[htb]
  \centering
  \includegraphics[width=\linewidth]{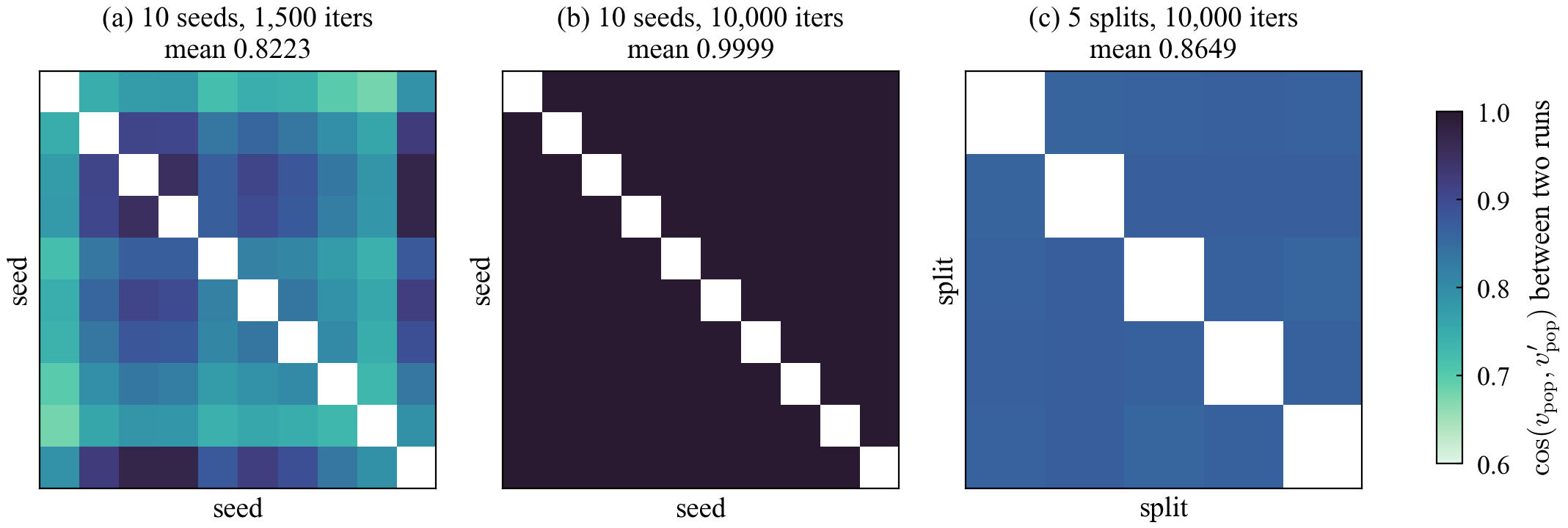}
  \caption{Pairwise cosine between $v_{\mathrm{pop}}$ fits. (a) Ten seeds at $1{,}500$
  iterations, where training has not converged and the spread has the appearance of several
  distinct solutions. (b) The same ten seeds at $10{,}000$ iterations, where the spread
  collapses to $0.9999$. (c) Five train/test splits at $10{,}000$ iterations, mean $0.8649$.}
  \label{fig:stability}
\end{figure}

\section{Compute}
\label{app:compute}

\paragraph{Embeddings.} Every experiment reads from one cached set of embeddings
per dataset, extracted in a single pass over the frozen
Skywork-Reward-Llama-3.1-8B-v0.2 backbone on one NVIDIA A10G. Apart from training the
sparse autoencoder below, this is the only stage that needs a GPU, and no
experiment re-runs it. The cached sets total
approximately $113{,}000$ forward passes: $48{,}385$ for Community Alignment,
$31{,}448$ for PRISM, $1{,}200$ for the synthetic response pool, and the
contrastive pairs behind the concept libraries. PRISM was extracted twice, so the
project total is higher than the caches imply. Extraction runs at batch size
$8$--$16$ on a single device. Across the project, all GPU work billed under
\$15 of A10G time on a serverless provider, on the order of ten to fifteen
GPU-hours at that instance's list price; this covers every extraction, including
the discarded PRISM pass and exploratory runs that are not reported.

\paragraph{Fitting.} Estimator fits run on CPU against those caches. One fit at
$K = 8$ over the 200 Community Alignment users takes about three minutes, so the
five-seed sweep across five ranks is roughly ninety minutes; the synthetic control
is smaller and faster. Memory is governed by the cache rather than the model,
at roughly $800$\,MB for Community Alignment and $650$\,MB for PRISM. The results
reported here rest on $301$ recorded fits across all sweeps, ranks and seeds,
which at the rate above is on the order of fifteen CPU-hours in total. No fit
requires a GPU, so the whole analysis is reproducible on a laptop once the
embedding caches exist.

\paragraph{Generation.} Contrastive text is written by Claude through a hosted
API. The eleven-concept library used for the alignment analysis comprises $550$
contrastive pairs generated by Claude Sonnet 4.5; the twelve-concept library and
the synthetic control both use Claude Opus 4.6, contributing $600$ contrastive
pairs and $1{,}200$ pool responses respectively. Generation is therefore on the
order of $2{,}400$ responses per library-and-pool build, small beside the
embedding pass and not repeated between experiments.

\paragraph{Sparse autoencoder.} The TopK autoencoder is trained on $62{,}984$
PRISM response activations, using both sides of every pair rather than their
difference. With $16{,}384$ features over $d = 4096$ it has $134$M parameters, and
trains for $20{,}000$ steps at batch size $256$ in approximately fifteen minutes on
a single A100-SXM4-40GB, under $4$\,GB of memory. Including the configurations it
was selected from and one retrain after the preprocessing correction, this is about
three GPU-hours in total.

\paragraph{Beyond the reported experiments.} The project used more compute than
the results above require. PRISM embeddings were extracted twice, the first pass
invalidated by the preprocessing defect of Appendix~\ref{app:reimpl}, and the
sweeps over $\alpha$ and over $\lambda_{\delta}$ ran wider than the subset
reported here.

%%%%%%%%%%%%%%%%%%%%%%%%%%%%%%%%%%%%%%%%%%%%%%%%%%%%%%%%%%%%

\newpage
\section*{NeurIPS Paper Checklist}

\begin{enumerate}

\item {\bf Claims}
    \item[] Question: Do the main claims made in the abstract and introduction accurately reflect the paper's contributions and scope?
    \item[] Answer: \answerYes{}
    \item[] Justification: Each contribution listed in the introduction maps to a corresponding result section: the RBD estimator (Section~\ref{sec:rbd}), the validation protocol (Sections~\ref{sec:synthetic-results} and~\ref{sec:perturbation}), the null per-user result on Community Alignment (Sections~\ref{sec:ca-ablation} and~\ref{sec:ca-unseen}) and on PRISM (Section~\ref{sec:perturbation}), and the generator-dependence of concept vectors (Section~\ref{sec:generator-dependence}). All results are obtained on two preference datasets and a single frozen reward backbone, and the paper claims no more.

\item {\bf Limitations}
    \item[] Question: Does the paper discuss the limitations of the work performed by the authors?
    \item[] Answer: \answerYes{}
    \item[] Justification: The Discussion includes a dedicated Limitations paragraph, covering the scope of the per-user recovery result and the restriction to two datasets and one reward backbone.

\item {\bf Theory assumptions and proofs}
    \item[] Question: For each theoretical result, does the paper provide the full set of assumptions and a complete (and correct) proof?
    \item[] Answer: \answerYes{}
    \item[] Justification: The paper's central formal claim is the invariance of the
objective under $V \mapsto VR$, $w \mapsto R^{-1}w$ for any invertible $R$, stated
in Section~\ref{sec:identifiability} and derived in
Appendix~\ref{app:identifiability} from the fact that the objective depends on the
parameters only through $V\bar{w}$ and $V\delta_u$. The paper accordingly reports
only quantities that are invariant under that transform. The one other formal
statement, that signed weights on a single column span the same reward functions
as non-negative weights on $\{v, -v\}$, is proved inline where it is used.

    \item {\bf Experimental result reproducibility}
    \item[] Question: Does the paper fully disclose all the information needed to reproduce the main experimental results of the paper to the extent that it affects the main claims and/or conclusions of the paper (regardless of whether the code and data are provided or not)?
    \item[] Answer: \answerYes{}
    \item[] Justification: Sections~\ref{sec:rbd} and~\ref{sec:setup} give the estimator, data, filters and splits, and Appendices~\ref{app:tuning},~\ref{app:synthetic} and~\ref{app:reimpl} give the tuning sweep, the construction of the synthetic control, and the corrections to the released LoRe code that the baseline depends on. Remaining details are in the released code.

\item {\bf Open access to data and code}
    \item[] Question: Does the paper provide open access to the data and code, with sufficient instructions to faithfully reproduce the main experimental results, as described in supplemental material?
    \item[] Answer: \answerYes{}
    \item[] Justification: Code and the synthetic control dataset are released at the anonymized repository linked in the abstract, with preparation and training commands for each dataset in the README. PRISM and Community Alignment are public datasets and are not redistributed, and the preference-pair embeddings and sparse-autoencoder checkpoint are regenerated by the released scripts rather than shipped.

\item {\bf Experimental setting/details}
    \item[] Question: Does the paper specify all the training and test details (e.g., data splits, hyperparameters, how they were chosen, type of optimizer) necessary to understand the results?
    \item[] Answer: \answerYes{}
    \item[] Justification: Section~\ref{sec:setup} gives the datasets, filters, backbone, splitting rules and the 20\%/20\% validation and test fractions, and Section~\ref{sec:rbd} the optimizer and the initialization scaling. Appendix~\ref{app:tuning} gives the tuning grids and how the hyperparameters were selected, and the configuration used for the interpretation experiments is stated in Section~\ref{sec:sae}.

\item {\bf Experiment statistical significance}
    \item[] Question: Does the paper report error bars suitably and correctly defined or other appropriate information about the statistical significance of the experiments?
    \item[] Answer: \answerYes{}
    \item[] Justification: Accuracy comparisons report mean $\pm$ standard deviation over seeds, with the seed count stated in each caption; the pooled identification figure of Figure~\ref{fig:identification} is a standard error instead, and the concept alignments of Figure~\ref{fig:concepts} are scored against a null of $10{,}000$ random directions at $p<0.05$. Tables~\ref{tab:sae} and~\ref{tab:perturbation-bipolar} report single fits or seed means without dispersion.

\item {\bf Experiments compute resources}
    \item[] Question: For each experiment, does the paper provide sufficient information on the computer resources (type of compute workers, memory, time of execution) needed to reproduce the experiments?
    \item[] Answer: \answerYes{} % Replace by \answerYes{}, \answerNo{}, or \answerNA{}.
    \item[] Justification:  Appendix~\ref{app:compute} reports the compute for each stage, including worker type, memory and running time.
    
\item {\bf Code of ethics}
    \item[] Question: Does the research conducted in the paper conform, in every respect, with the NeurIPS Code of Ethics \url{https://neurips.cc/public/EthicsGuidelines}?
    \item[] Answer: \answerYes{}
    \item[] Justification: The work conforms to the NeurIPS Code of Ethics: it analyses two publicly released preference datasets, collects no new human data, releases no model weights, and preserves anonymity in the linked repository.

\item {\bf Broader impacts}
    \item[] Question: Does the paper discuss both potential positive societal impacts and negative societal impacts of the work performed?
    \item[] Answer: \answerYes{}{} % Replace by \answerYes{}, \answerNo{}, or \answerNA{}.
    \item[] Justification: Discussion section outlines the broader impacts of the work.
    
\item {\bf Safeguards}
    \item[] Question: Does the paper describe safeguards that have been put in place for responsible release of data or models that have a high risk for misuse (e.g., pre-trained language models, image generators, or scraped datasets)?
    \item[] Answer: \answerNA{}
    \item[] Justification: The release is analysis code, a synthetic preference dataset generated for the purpose, and the low-dimensional directions and basis matrices used in the analyses. It includes no language model, image generator, or scraped data, and the reward backbone is an existing public checkpoint that is not redistributed.

\item {\bf Licenses for existing assets}
    \item[] Question: Are the creators or original owners of assets (e.g., code, data, models), used in the paper, properly credited and are the license and terms of use explicitly mentioned and properly respected?
    \item[] Answer: \answerYes{}
    \item[] Justification: Every asset is credited by citation at first use. PRISM is released under CC-BY-4.0 for human-written prompts and CC-BY-NC-4.0 for model responses, Community Alignment under CC-BY-4.0, the Skywork-Reward-Llama-3.1-8B-v0.2 backbone under the Skywork Community License over its Llama 3.1 base terms, and the LoRe implementation we fork under CC-BY-NC 4.0; all are used non-commercially for research, and our released code retains the LoRe license.

\item {\bf New assets}
    \item[] Question: Are new assets introduced in the paper well documented and is the documentation provided alongside the assets?
    \item[] Answer: \answerYes{}
    \item[] Justification: The new assets are the synthetic control dataset ($60$ personas, $1{,}200$ responses, $14{,}400$ preference pairs), the RBD and sparse-autoencoder implementations, and the fitted directions, all released in the anonymized repository under the CC-BY-NC 4.0 license inherited from the LoRe fork. Appendix~\ref{app:synthetic} documents the dataset's construction.

\item {\bf Crowdsourcing and research with human subjects}
    \item[] Question: For crowdsourcing experiments and research with human subjects, does the paper include the full text of instructions given to participants and screenshots, if applicable, as well as details about compensation (if any)? 
    \item[] Answer: \answerNA{}
    \item[] Justification: The paper conducts no crowdsourcing and collects no data from human subjects. It analyses two preference datasets, PRISM and Community Alignment, that were collected and publicly released by their original authors, and all remaining preference labels are synthetic.

\item {\bf Institutional review board (IRB) approvals or equivalent for research with human subjects}
    \item[] Question: Does the paper describe potential risks incurred by study participants, whether such risks were disclosed to the subjects, and whether Institutional Review Board (IRB) approvals (or an equivalent approval/review based on the requirements of your country or institution) were obtained?
    \item[] Answer: \answerNA{}
    \item[] Justification: No human subjects research was conducted for this paper. It analyses two publicly released datasets, involves no participant contact, and incurs no additional risk to the people who contributed to them.

\item {\bf Declaration of LLM usage}
    \item[] Question: Does the paper describe the usage of LLMs if it is an important, original, or non-standard component of the core methods in this research? Note that if the LLM is used only for writing, editing, or formatting purposes and does \emph{not} impact the core methodology, scientific rigor, or originality of the research, declaration is not required.
    %this research? 
    \item[] Answer: \answerYes{}
    \item[] Justification: LLMs generate the contrastive text that the concept libraries and the synthetic control are built from, so they are part of the core methodology: Claude Sonnet 4.5 wrote the eleven-concept library of Section~\ref{sec:population}, and Claude Opus 4.6 the twelve-concept library and persona response pool of Appendix~\ref{app:synthetic}. An LLM accept/reject judge additionally filters each contrastive pair during generation, and Section~\ref{sec:generator-dependence} measures the resulting generator dependence.

\end{enumerate}

\end{document}